%% file: main.tex
\documentclass[10pt,twocolumn,letterpaper]{article}

 \usepackage{cvpr}              % To produce the CAMERA-READY version
\input{preamble}
\definecolor{cvprblue}{rgb}{0.21,0.49,0.74}
\usepackage[pagebackref,breaklinks,colorlinks,allcolors=cvprblue]{hyperref}

\def\ie{\textit{i.e.}}

\def\etc{\textit{etc.}}

\def\paperID{9214} % *** Enter the Paper ID here
\def\confName{CVPR}
\def\confYear{2026}

\title{S\textsuperscript{2}AM3D: Scale-controllable Part Segmentation of 3D Point Clouds}

\author{
Han Su$^{1}$ \quad
Tianyu Huang$^{1}$ \quad
Zichen Wan$^{1}$ \quad
XiaoHe Wu$^{1}$ \quad
Wangmeng Zuo$^{1}$ \footnotemark[1] \\
$^{1}$Harbin Institute of Technology \\
}

\begin{document}
\maketitle
\input{sec/0_abstract}    
\input{sec/1_intro}
\input{sec/2_related_work}
\input{sec/3_method}

\input{sec/4_data}
\input{sec/5_experiment}

\input{sec/6_conclusion}
{
    \small
    \bibliographystyle{ieeenat_fullname}
    \bibliography{main}
}


\end{document}

% \maketitle
\maketitlesupplementary

This supplementary material presents additional implementation details (Section~\ref{sec:imple_detail}), additional experiments (Section~\ref{sec:addition}), and more visualizations (Section~\ref{sec:more}).

\section{Implementation Details}\label{sec:imple_detail}
\noindent\textbf{Decoder.}
The continuous scale $s$ is encoded with $M=64$ sinusoidal frequency pairs and passed through a LayerNorm followed by a linear layer to produce the channel-wise FiLM~\cite{perez2018film} parameters $(\boldsymbol{\gamma},\boldsymbol{\beta})$.
The linear layer is initialized with zero weights and biases, so the FiLM branch starts with an identity transformation.
We additionally use a global scaling factor $\alpha$ to modulate the FiLM output, which is initialized to $0.1$.

\noindent\textbf{Data pipeline.} We train a PointNet-based~\cite{qi2017pointnet++} validator on a manually selected subset of 800 shapes (400 valid and 400 invalid), which is split into 70\%/30\% training and test sets. To rigorously exclude unqualified samples, we use a confidence threshold of 0.8, achieving a Precision/Recall of 1.00/0.78 on the test set. To mitigate category bias during validator training, we construct the manually labeled dataset via random, class-balanced sampling. The filtering criterion is primarily based on label distributions and is only weakly correlated with geometric structure, reducing the risk of shape-specific bias. We manually inspected approximately 10\% of the final 100k-scale dataset, confirming the high quality and broad diversity of the resulting dataset. Overall, the filtering process retains about 60\% of the shapes. For connectivity refinement, we apply DBSCAN~\cite{ester1996density} with radius $\varepsilon = d \times \varepsilon_{\text{factor}}$ using $\varepsilon_{\text{factor}}=0.15$, where $d$ is the diagonal length of the axis-aligned bounding box of the label, to split spatially disconnected components into distinct parts.

\noindent\textbf{Full segmentation.} We perform full segmentation in a simulated interactive setting. For each ground-truth part, we select a prompt point and use the same scale definition as in interactive segmentation. The model generates one candidate mask per part using the same configuration as in the interactive experiments. We then apply lightweight post-processing (see Algorithm~\ref{alg:postproc}): overlapping points are assigned to masks based on prediction confidence and distance to the mask centers, followed by $k$-NN label propagation to cover the remaining points, producing a full segmentation.
\begin{figure}[t]
    \centering
    \includegraphics[width=0.85\linewidth]{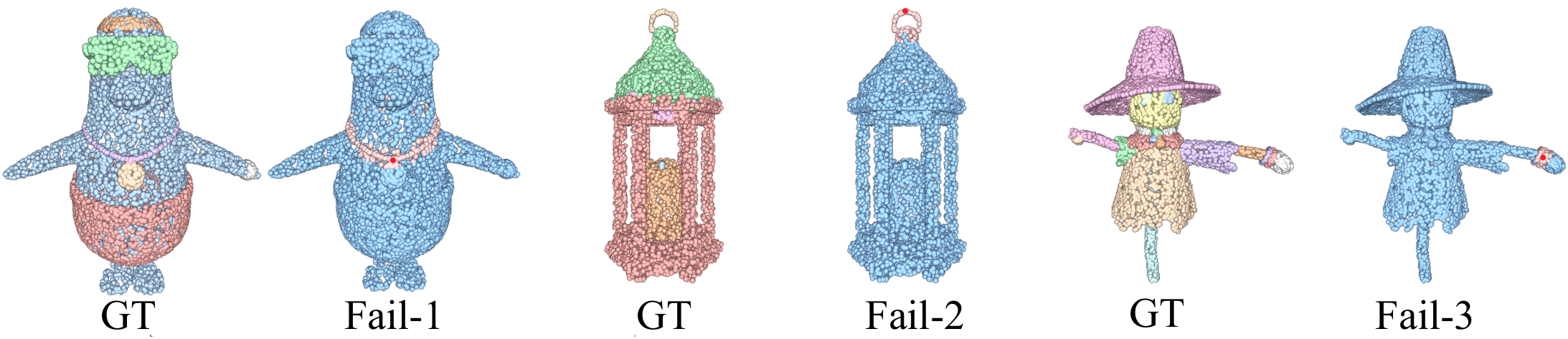}
    \caption{Representative failure cases of S$^2$AM3D in challenging scenarios.}
    \label{fig:failure_cases_suppl}
\end{figure}
\begin{algorithm}[t]
    \centering
    \small
    \caption{Post-Processing for Full Segmentation}
    \label{alg:postproc}
    \begin{algorithmic}[1]
        \STATE \textbf{Input}: point cloud $\mathbf{X}$, candidate masks $\{\mathbf{m}_k\}$, point-wise scores $\{p_{ik}\}$, confidence–distance weight $\alpha$, $k$-NN size $k$
        \STATE \textbf{Output}: final disjoint masks $\{\tilde{\mathbf{m}}_j\}$

        \STATE $\blacktriangleright$ \textbf{Overlap Resolution}
        \STATE Compute geometric centers $\boldsymbol{\mu}_k$ for each candidate mask $\mathbf{m}_k$
        \FOR{each point $i$}
            \STATE Let $\mathcal{J}_i = \{k \mid \mathbf{m}_k(i)=1\}$ be the set of masks covering point $i$
            \IF{$|\mathcal{J}_i| = 1$}
                \STATE Assign point $i$ to that single mask
            \ELSIF{$|\mathcal{J}_i| > 1$}
                \STATE For each $k \in \mathcal{J}_i$, compute
                \[
                    d_{ik} = \|\mathbf{X}_i - \boldsymbol{\mu}_k\|_2,\quad
                    s^{\text{conf}}_{ik} = p_{ik},\quad
                    s^{\text{dist}}_{ik} = \exp(-d_{ik})
                \]
                \STATE Combine them as
                \[
                    \text{score}_{ik} = \alpha\, s^{\text{conf}}_{ik} + (1-\alpha)\, s^{\text{dist}}_{ik}
                \]
                \STATE Assign point $i$ to the mask $k$ with the highest $\text{score}_{ik}$
            \ENDIF
        \ENDFOR

        \STATE $\blacktriangleright$ \textbf{$k$-NN Label Propagation}
        \STATE Build a $k$-NN graph over $\mathbf{X}$
        \FOR{$t = 1$ \TO $5$}
            \FOR{each point not assigned to any mask}
                \STATE Collect mask assignments of its $k$ nearest neighbors and update its assignment by majority vote
            \ENDFOR
        \ENDFOR
        \STATE Optionally, for each $i$ with $z_i = -1$, set $z_i \leftarrow z_{k^\star}$ where 
       $k^\star = \arg\min_{k:\,z_k \neq -1} \|\mathbf{x}_i - \mathbf{x}_k\|_2$
        \STATE Return disjoint masks $\{\tilde{\mathbf{m}}_j\}$ with $\tilde{\mathbf{m}}_j(i) = \mathbb{1}[z_i = j]$

        \RETURN $\{\tilde{\mathbf{m}}_j\}$
    \end{algorithmic}
\end{algorithm}

\input{tab/suppl_density_tab}
\input{tab/suppl_complex_tab}
\input{tab/suppl_scale_tab}

\section{Additional Experiments}\label{sec:addition}

\subsection{Density Analysis}
Table~\ref{tab:density_robust} presents the density analysis, reporting IoU at 10k and 100k input points on PartObjaverse-Tiny~\cite{yang2024sampart3d} and PartNet-E~\cite{liu2023partslip}. The results show consistent performance at both densities: the impact on PartObjaverse-Tiny~\cite{yang2024sampart3d} is minimal, while higher density brings moderate gains on PartNet-E~\cite{liu2023partslip}. These results indicate that S$^2$AM3D is robust to variations in point density in practical settings.

\subsection{Complexity Analysis}

Table~\ref{tab:complexity} presents the complexity analysis, reporting the inference time per point prompt in the interactive setting and the number of network parameters, measured on a single NVIDIA H20 GPU.
\subsection{Scale Analysis}
In the main experiments, the scale prompt $s$ is set to the ground-truth part size for controlled evaluation. In practice, $s$ may instead be provided by a user's rough estimate or by external modules. To assess sensitivity, we perturb the scale by $s'=(1+\delta)s$ and measure the IoU change relative to $\delta=0$. As shown in Table~\ref{tab:scale_robustness}, on PartObjaverse-Tiny~\cite{yang2024sampart3d}, perturbations within $\pm20\%$ cause negligible performance change ($|\Delta\text{IoU}|<1.0$). For moderate perturbations ($|\delta|\in 30\%\text{--}50\%$), IoU decreases slowly, indicating gradual rather than catastrophic degradation. For larger perturbations ($|\delta|\geq 100\%$), the perturbed scale effectively corresponds to a different semantic granularity from the original annotation; in this range ($100\%\text{--}300\%$), the IoU drop is more pronounced, which is expected when still evaluating against the original ground-truth part.

\subsection{Failure Analysis}
Due to the frequent absence and low quality of textures in Objaverse, our training and evaluation are based on XYZ input. We visualize failure cases in Fig.~\ref{fig:failure_cases_suppl}, which mainly occur on extremely thin/long structures (Fail-1,2) or on geometry-similar parts that require color cues to distinguish (Fail-3, two wristbands merging together).
\input{fig/more_vis_fig}

\section{More Visualization Results}\label{sec:more}
Additional qualitative results of S$^2$AM3D for full segmentation and interactive segmentation are shown in Figure~\ref{fig:more_vis}. The method produces accurate and fine-grained part predictions across diverse object categories and geometric structures.

{
    \small
    \bibliographystyle{ieeenat_fullname}
    \bibliography{main}
}


%% file: sec/0_abstract.tex
\begin{abstract}

Part-level point cloud segmentation has recently attracted significant attention in 3D computer vision.
Nevertheless, existing research is constrained by two major challenges: native 3D models lack generalization due to data scarcity, while introducing 2D pre-trained knowledge often leads to inconsistent segmentation results across different views.
To address these challenges, we propose S$^2$AM3D, which incorporates 2D segmentation priors with 3D consistent supervision. We design a point-consistent part encoder that aggregates multi-view 2D features through native 3D contrastive learning, producing globally consistent point features. A scale-aware prompt decoder is then proposed to enable real-time adjustment of segmentation granularity via continuous scale signals. Simultaneously, we introduce a large-scale, high-quality part-level point cloud dataset with more than 100k samples, providing ample supervision signals for model training.
Extensive experiments demonstrate that S$^2$AM3D achieves leading performance across multiple evaluation settings, exhibiting exceptional robustness and controllability when handling complex structures and parts with significant size variations.
The project page is available at \url{https://sumuru789.github.io/S2AM3D-website/}.
\end{abstract}

%% file: sec/1_intro.tex
\section{Introduction}

Part-level point cloud segmentation plays a pivotal role in bridging fine-grained geometric details with high-level semantic understanding, supporting significant applications in 3D content creation, robotic manipulation, and reverse engineering~\cite{zhang2025physchoreophysicscontrollablevideogeneration,chen2025partgen,geng2023gapartnet,liu2024point2cad,liu2024comprehensive,huang2024dreamcontrol}, \etc
 Unlike instance detection, which only provides holistic semantics, 
part-level segmentation enables flexible granularity adjustment and seamless switching between local and global regions. This capability directly influences the feasibility and efficiency of downstream tasks, including part generation, replacement, assembly, and parametric editing.
However, high-quality 3D part-level data remains scarce. 
Compared with the 2D domain, the cost of point cloud annotation is more prohibitive; consequently, existing datasets suffer from limited scale and category diversity, which severely hinders the generalization performance of 3D models~\cite{qi2017pointnet,qi2017pointnet++,wang2018sgpn,li2018pointcnn,ma2025p3}.

\input{fig/intro_demo}

Recent works~\cite{liu2023partslip,yang2024sampart3d,liu2025partfield,zhou2024point,zhou2023partslip++,garosi20253d,Sun_2024_CVPR} leverage 2D pre-trained knowledge to enhance the generalization of 3D models, \ie, adopting 2D segmentation on 3D renderings. Segmented multi-view contents are then aggregated via mechanisms like multi-view lifting~\cite{garosi20253d,Sun_2024_CVPR,liu2023partslip,zhou2023partslip++} or distillation~\cite{yang2024sampart3d,liu2025partfield,zhou2024point} to obtain the final 3D segmentation results.
However, cross-view inconsistencies caused by occlusions, slender structures, and complex topology, 
can lead to accumulated errors that compromise global 3D consistency.
To alleviate the inconsistency, some methods~\cite{liu2025partfield,zhou2024point} further incorporate 2D priors and provide supervision in the 3D space.
Although native 3D data is used to enhance multi-view consistency, these methods still rely too heavily on 2D segmentation results.
For example, PartField~\cite{liu2025partfield} primarily leverages part proposals from SAM~\cite{kirillov2023segment}, supplemented with limited 3D supervision to learn multi-view dependencies.
Consequently, when 2D segmentation is affected by occlusions and other spatial factors, 3D representations struggle to correct these issues (see Figure~\ref{fig:intro_demo}).

In contrast to the 2D domain, point cloud segmentation is more challenging, considering its requirements of integrating both global and local context. The absence of global information can compromise multi-view consistency, while an excessive focus on global features may cause distortions in the correlation between local parts.
To better combine the spatial features with 2D priors, we introduce \textbf{S\textsuperscript{2}AM3D}, a multi-modal joint-supervised framework for scale-aware point cloud part segmentation. 
We propose a point-consistent part encoder to aggregate global information from multi-view 2D features with contrastive supervision of point cloud data. Additionally, given the inherent weakness of global representations in encoding the relationships of local parts, we propose a scale-aware prompt decoder to explicitly model cross-granularity relations. To this end, the input scale is mapped to a learnable sinusoidal embedding and then fed to bi-directional cross-attention modules together with encoded global point features. Using the prompt point as a query, the attention layers fuse multi-granularity context to finally determine the per-point probabilities in one pass.

Currently, existing part-level point cloud datasets generally suffer from limited scale and insufficient annotation fidelity. To support our point-consistent 3D supervision, we introduce an automatic data pipeline for scalable part-level point cloud data curation, including part annotation, quality filtering, and connectivity refinement. Filtering and refinement strictly enforce 3D geometric constraints on the raw annotations, ensuring the quality of processed labels. Using this pipeline, we collect a dataset of more than 100,000 point cloud instances across 400 categories, annotated with about 1.2 million fine-grained part labels.
%To the best of our knowledge, this is the largest available 3D part dataset to date~\cite{yang2024sampart3d,liu2023partslip,mo2019partnet,wang2025partnext,xiang2020sapien,yi2016scalable,chang2015shapenet} (see Figure~\ref{fig:intro_demo}); we will release it publicly.

Extensive experiments are conducted to demonstrate the effectiveness of S\textsuperscript{2}AM3D on part-level point cloud segmentation. 
%Experimental results show that our model outperforms state-of-the-art methods, with a substantial margin on both PartObjaverse-Tiny~\cite{yang2024sampart3d} and PartNet-E~\cite{liu2023partslip} benchmarks. 
% {\color{blue}On both PartObjaverse-Tiny~\cite{yang2024sampart3d} and PartNet-E~\cite{liu2023partslip} benchmarks, our 2D–3D training recipe enables S\textsuperscript{2}AM3D to achieve performance comparable to P\textsuperscript{3}-SAM~\cite{ma2025p3} while using far less training data; and significantly outperforms SAMPart3D and PartField when using a comparable amount of data. (see Figure~\ref{fig:intro_demo})}
As shown in Figure~\ref{fig:intro_demo}, our 2D-3D hybrid training recipe allows S\textsuperscript{2}AM3D to achieve comparable performance to P\textsuperscript{3}-SAM~\cite{ma2025p3} with far less training data, and significantly outperforms SAMPart3D and PartField.
Benefitting from the proposed scale-aware prompt decoder, S\textsuperscript{2}AM3D exhibits remarkable flexibility by naturally extending its capability to point-prompted segmentation, making it a unified 3D part segmentation framework.

Our contributions can be summarized as:
\begin{itemize}
\item We propose a 2D-3D training recipe for part segmentation: it reuses 2D pre-trained knowledge and conducts native 3D supervision to yield globally point-consistent part features.
\item We propose a scale-aware prompt decoder with a scale modulator and bi-directional cross-attention, enabling flexible 3D part segmentation.
\item We introduce a scalable data pipeline for part-level point cloud data curation, collecting over 100,000 labeled point cloud instances for training.
\end{itemize}

%% file: fig/intro_demo.tex
\begin{figure}[t]
    \centering
    \includegraphics[width=0.47\textwidth]{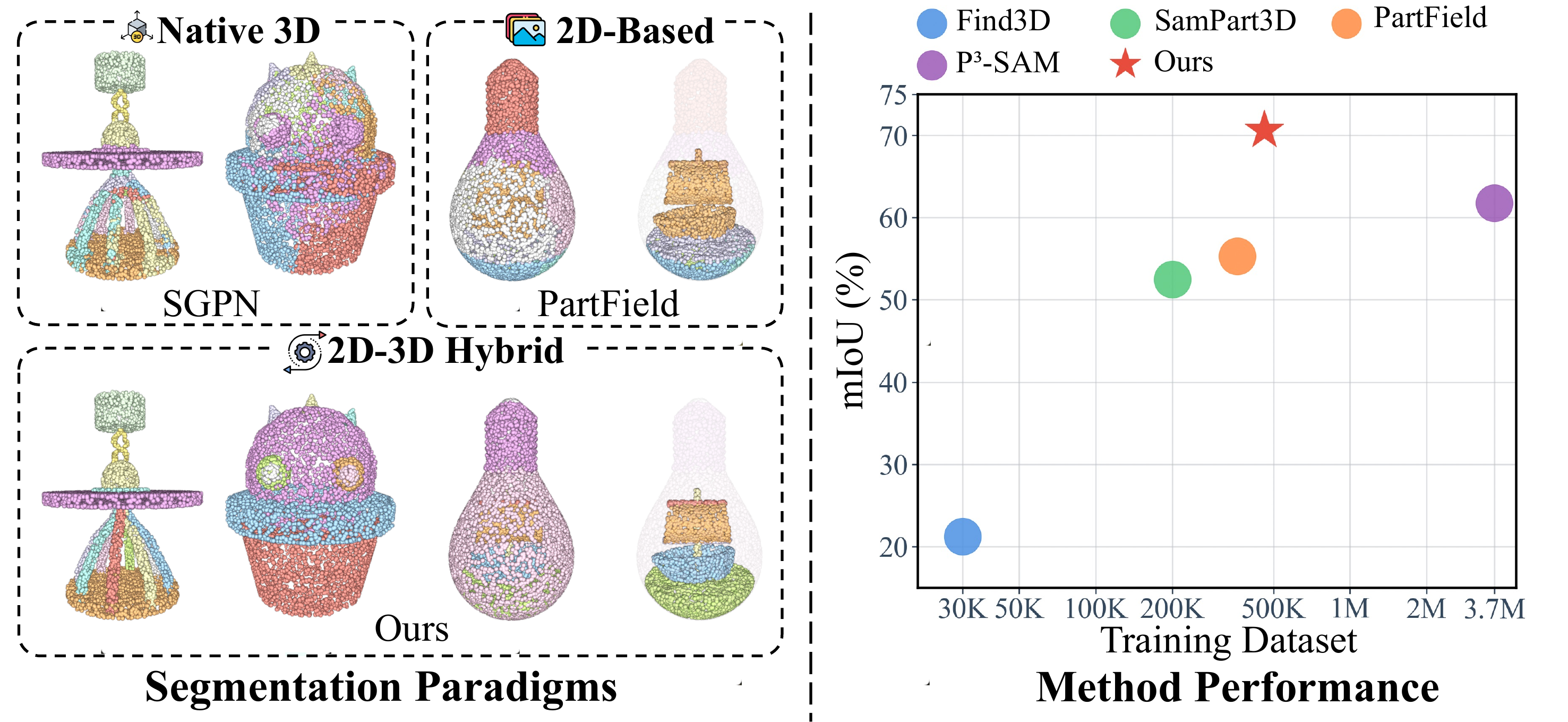}
    % \caption{SGPN~\cite{wang2018sgpn} shows limited generalization; PartField~\cite{liu2025partfield} exhibits cross-view inconsistencies and is sensitive to occlusions and complex topology (left). Comparison of dataset statistics across open source 3D part segmentation datasets (right)~\cite{mo2019partnet,yi2016scalable,chang2015shapenet,yang2024sampart3d,liu2023partslip,wang2025partnext,xiang2020sapien}.}
    \caption{Paradigm comparison (left): Native 3D methods present limited generalization, and 2D-based methods fail in complex cases like occlusions. Our hybrid solution solves these issues. Performance Comparison (right): Our method reaches large-scale training performance with much less data and significantly outperforms previous methods at similar data scales.}
    \label{fig:intro_demo}
\end{figure}

%% file: sec/2_related_work.tex
\section{Related Work}

\noindent\textbf{Native 3D Part Segmentation.}
Early point cloud/mesh segmentation methods were typically trained under closed-label settings on ShapeNet-Part and PartNet, leveraging pointwise or neighborhood aggregation networks such as PointNet/PointNet++ and the Point Transformer family~\cite{qi2017pointnet,qi2017pointnet++,zhao2021point,yi2016scalable,mo2019partnet}. While strong within restricted categories and fixed granularities, these models often struggle under open-domain shapes, long-tailed parts, and cross-source shifts: (i) limited category/hierarchy coverage hampers generalization to unseen objects and rare parts; (ii) geometric-detail and noise discrepancies across sources obscure unified supervision; and (iii) the absence of explicit hierarchical consistency and granularity control constrains editing and interactive scenarios~\cite{chang2015shapenet,mo2019partnet,collins2022abo}. These limitations motivate scalable part understanding beyond closed-world assumptions.

\noindent\textbf{2D Transferred Part Segmentation.}
To mitigate the scarcity of native 3D annotations, many works transfer 2D foundation priors (e.g., SAM, CLIP/GLIP) to 3D~\cite{kirillov2023segment,radford2021learning,li2022grounded}. Two mainstream paradigms emerge: (i) \emph{multi-view lifting} renders shapes to images, applies 2D segmentation or vision--language priors per view, and fuses/back-projects/distills view-level cues into 3D supervision~\cite{kirillov2023segment,yang2023sam3d,xu2023sampro3d,liu2023partslip}; and (ii) \emph{2D-to-3D distillation} directly trains native 3D networks from image-domain pseudo labels~\cite{huang2024segment3d,liu2025partfield,yang2024sampart3d}. Despite reduced labeling cost and open-vocabulary semantics, both paradigms are prone to cross-view inconsistencies on thin/occluded structures, boundary ambiguity that undermines global 3D coherence, and costly post-processing pipelines; moreover, surface-only coverage limits supervision of internal structures~\cite{yang2023sam3d,xu2023sampro3d,liu2023partslip,huang2024segment3d}. These observations point to the value of native 3D signals for globally consistent part supervision.

\noindent\textbf{Granularity Controllable Part Segmentation.}
Existing methods still face significant limitations in achieving flexible control over segmentation granularity. Feature clustering-based approaches (e.g., PartField~\cite{liu2025partfield} and SAMPart3D~\cite{yang2024sampart3d}) typically rely on post-processing clustering to determine segmentation granularity, resulting in non-continuous and non-intuitive control that hinders real-time fine-grained adjustment. Point prompt-based methods (e.g., Point-SAM~\cite{zhou2024point} and P\textsuperscript{3}-SAM~\cite{ma2025p3}), while supporting interactive point prompting, lack explicit granularity control mechanisms and cannot precisely regulate segmentation scope through continuous parameters. Furthermore, the segmentation granularity in both types of methods remains fundamentally constrained by their reliance on 2D priors, making it challenging to ensure global consistency in 3D space. These limitations motivate us to explore a new solution that introduces an explicit continuous scale signal coupled with native 3D supervision, ultimately achieving truly reliable and controllable 3D segmentation granularity.

\input{fig/pipeline_fig}

%% file: fig/pipeline_fig.tex
\begin{figure*}[t]
    \centering
    \includegraphics[width=\textwidth]{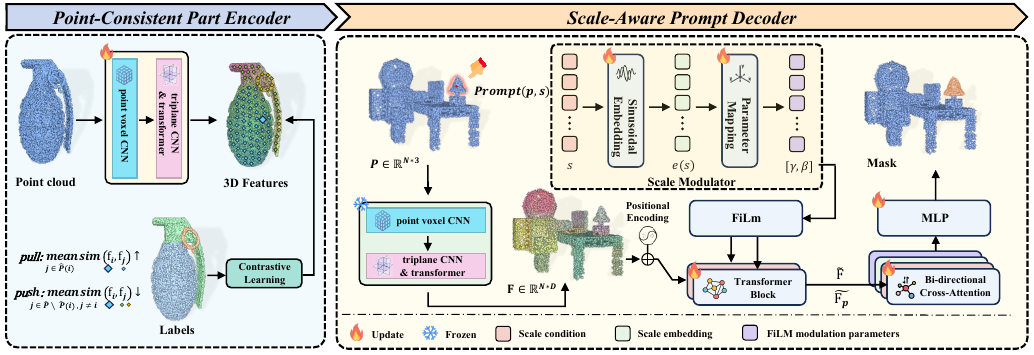}
\caption{S\textsuperscript{2}AM3D pipeline. Left: under 3D supervision with contrastive learning, the input point cloud $\mathbf{P}\in\mathbb{R}^{N\times3}$ is encoded into per-point features $\mathbf{F}\in\mathbb{R}^{N\times D}$. Right: given a prompt $(p,s)$, $s$ is mapped by a sinusoidal embedding $\mathbf{e}(s)$ to FiLM parameters $[\gamma,\beta]$, which perform channel-wise modulation to obtain a scale-enhanced representation $\tilde{\mathbf{F}}$; the prompt vector $\tilde{\mathbf{F}}_{p}$ is then indexed and interacts with the global features via bi-directional cross-attention, after which an MLP and a Sigmoid produce a probability mask.}

\vspace{-1em}
    \label{fig:pipeline}
\end{figure*}

%% file: sec/3_method.tex
\section{Method}

\subsection{Overview}\label{sec:overview}
We propose \textbf{S\textsuperscript{2}AM3D}, a \emph{scale-controllable}, \emph{point-prompted} framework for part segmentation of 3D point clouds. 
Given the input point cloud $P\!\in\!\mathbb{R}^{N\times3}$, we first extract point-consistent features by incorporating 2D segmentation priors with 3D contrastive supervision (Sec.~\ref{sec:encoder}). 
Conditioned on a point prompt index $p$ with an optional scale prompt $s\in[0,1]$, our Scale-Aware Prompt Decoder enables flexible part segmentation via scale modulator and bi-directional cross-attention (Sec.~\ref{sec:decoder}). 
We adopt a decoupled training scheme, \ie, first stabilizing the encoder representation, then training the scale-aware decoder on top of the frozen encoder (Sec.~\ref{sec:training}). The overall framework is shown in Figure~\ref{fig:pipeline}.

\subsection{Point-Consistent Part Encoder}\label{sec:encoder}

To aggregate point cloud features with general segmentation priors, the common practice is to train a point cloud encoder by distilling the pre-training knowledge from 2D segmentation models like SAM~\cite{kirillov2023segment}. Following PartField ~\cite{liu2025partfield}, we exploit a voxel-based encoder PVCNN~\cite{liu2019point} to extract the point latent and then convert it to the tri-plane~\cite{chan2022efficient} ($xy$, $yz$, and $zx$) representation with orthogonal projection. The Tri-plane features $\mathbf{T}\!\in\!\mathbb{R}^{3\times D \times H \times W}$ are then aggregated by a series of Transformer blocks, where $D$ is the feature channel. During training, tri-plane features are rendered from random viewpoints to formulate a 2D latent, which can be supervised with the distillation of SAM. However, relying solely on multi-view 2D distillation tends to introduce multi-view inconsistencies, leading to segmentation problems like boundary artifacts and cross-view conflicts.

Therefore, we additionally perform a \emph{native 3D contrastive supervision} under labeled point cloud data to enhance global consistency. 
To operationalize this objective in the construction of contrastive pairs, we restrict contrasts to \emph{intra-instance}, \ie, each mini\mbox{-}batch contains a single object. This mechanism ensures that positive and negative points are drawn from the same instance to avoid cross-instance semantic mismatches.
We randomly subsample points from a single instance and denote their indices by $\hat{P}$ (with $|\hat{P}|=\hat{N}$).
For any anchor $i\in\hat{P}$ with label $y_i$, the positive set is:
\[
\hat{P}(i)=\{\,j\in\hat{P}\setminus\{i\}\mid y_j=y_i\,\}.
\]

We use cosine similarity with temperature $\tau$ between $\ell_2$-normalized features $\mathbf{f}_i$ and $\mathbf{f}_j$:
$s_{ij}=\mathbf{f}_i^\top\mathbf{f}_j/\tau.$
The contrastive objective is
\[
\mathcal{L}_{\text{contr}}
= \frac{1}{|\hat{P}|}\sum_{i\in\hat{P}}
-\log
\frac{\sum_{j\in \hat{P}(i)} e^{\,s_{ij}}}
{\sum_{j\in \hat{P}\setminus\{i\}} e^{\,s_{ij}}}.
\]

To obtain point features $\mathbf{F}$ given the 3D coordinate $(x,y,z)$ from the tri-plane representation~\cite{chan2022efficient}, we unproject the point to the three feature planes of $\mathbf{T}$. The sampled features are summed across planes and stacked over all points as:
\begin{equation}
\mathbf{F}
=\Big[\,\mathbf{T}_{xy}(x_n,y_n)+\mathbf{T}_{yz}(y_n,z_n)+\mathbf{T}_{zx}(z_n,x_n)\,\Big]_{n=1}^{N}
\end{equation}

This objective compacts intra-part clusters and enlarges inter-part margins, yielding globally coherent per-point embeddings and sharper boundaries for the subsequent scale-aware prompt decoding.

\subsection{Scale-Aware Prompt Decoder}\label{sec:decoder}
Given point features $\mathbf{F}\in\mathbb{R}^{N\times D}$ and 3D coordinates $\mathbf{P}\in\mathbb{R}^{N\times 3}$, we compute a 3D sinusoidal positional encoding $\mathrm{PE}(\mathbf{P})$ and obtain the base representation
\begin{equation}
\mathbf{X}^{(0)}=\mathbf{F}+\mathrm{PE}(\mathbf{P}).
\end{equation}
However, such global representations often struggle to capture fine-grained intra-class multi-scale variations and local relations among parts, whereas interactive applications require continuously controllable granularity from fine to coarse. 
Therefore, beyond the point prompt, we further introduce the \emph{scale} prompting to guide feature decoding and adopt bi-directional cross-attention to enhance the point prompting.

\noindent\textbf{Scale Modulator.}
The scale $s$ is defined by the part’s relative size, which is the ratio of its points to the total number of points. Scale information should influence the global representation in a \emph{cross-layer transferable} manner.
For a continuous scale $s\in[0,1]$, we construct a learnable sinusoidal embedding
\begin{equation}
\mathbf{e}(s)=\big[\sin(\omega_k s+\phi_k),\ \cos(\omega_k s+\phi_k)\big]_{k=1}^{M},
\end{equation}
where $\{\omega_k,\phi_k\}$ are learnable and $M$ denotes the number of sinusoidal frequency pairs. 
$\mathbf{e}(s)$ is then fed to a feature-wise linear modulation~\cite{perez2018film} (FiLM) in the channel dimension, which modulates the global feature map with a learnable gate $\alpha$:
\begin{equation}
\begin{aligned}
\left[\boldsymbol{\gamma}, \boldsymbol{\beta}\right] &= \text{Linear}\!\big(\mathrm{LN}(\mathbf{e}(s))\big), \\
\mathrm{FiLM}(\mathbf{X};s) &= \mathbf{X} \odot \left(1 + \alpha \boldsymbol{\gamma}\right) + \alpha \boldsymbol{\beta}.
\end{aligned}
\end{equation}
where Linear() is a linear layer applied after Layer Normalization on $\mathbf{e}(s)$, 
$\boldsymbol{\gamma}, \boldsymbol{\beta} \in \mathbb{R}^D$ are channel-wise FiLM parameters~\cite{perez2018film},
and $\alpha \in \mathbb{R}$ is a learnable scalar gate.
To maintain a consistent scale context across semantic levels, we interleave FiLM with Transformer blocks $T$:
\begin{equation}
\mathbf{X}^{(\ell+1)} = T_\ell\!\big(\mathrm{FiLM}(\mathbf{X}^{(\ell)};s)\big),\qquad \ell=0,\dots,L_m-1.
\end{equation}
The resulting scale-conditioned enhanced representation is
\begin{equation}
\tilde{\mathbf{F}}=\mathbf{X}^{(L_m)}.
\end{equation}

To support inference without scale input, we randomly drop out the scale as $\mathbf{e}(s)=\mathbf{0}$ during training. FiLM naturally degenerates to the identity, i.e., $\mathrm{FiLM}(\mathbf{X};0)=\mathbf{X}$.

\noindent\textbf{Bi-directional Cross-Attention.} Unidirectional cross-attention struggles to perform both context aggregation and fine-grained refinement in one pass. To address this, we adopt \emph{bi-directional} cross-attention for joint localization and refinement.
We have scale-aware features $\tilde{\mathbf{F}}\in\mathbb{R}^{N\times D}$ and the corresponding feature of the point prompt $\tilde{\mathbf{F}_p}\in\mathbb{R}^{1\times D}$.
Let the input to the $\ell$-th layer be $(\mathbf{Y}^{(\ell)}, \mathbf{q}^{(\ell)})$, with $\mathbf{Y}^{(0)}=\tilde{\mathbf{F}}$ and $\mathbf{q}^{(0)}=\tilde{\mathbf{F}_p}$. Each layer comprises two cross-attention steps with explicit Q/K/V roles, followed by an FFN:
\begin{equation}
\label{eq:bi-ca-update}
\begin{aligned}
\mathrm{CAttn}(A;B) &= \mathrm{MHA}(Q{=}A,\,K{=}B,\,V{=}B),\\
\mathbf{q}^{(\ell+1)} &= \mathbf{q}^{(\ell)} + \mathrm{CAttn}\!\big(\mathbf{q}^{(\ell)};\,\mathbf{Y}^{(\ell)}\big),\\
\mathbf{Y}^{(\ell+1)} &= \mathrm{FFN}\!\Big(\mathbf{Y}^{(\ell)} + \mathrm{CAttn}\!\big(\mathbf{Y}^{(\ell)};\,\mathbf{q}^{(\ell+1)}\big)\Big),
\end{aligned}
\end{equation}
\noindent\textit{where } $\mathrm{MHA}$ denotes \textit{MultiHeadAttention}, and $\ell=0,\dots,L_d-1$.
 After stacking $L_d$ layers, we obtain interaction-enhanced features $\mathbf{H}=\mathbf{Y}^{(L_d)}$, and a lightweight segmentation head outputs pointwise probabilities:
\begin{equation}
\mathbf{o}=\mathrm{MLP}(\mathbf{H}),\qquad
\hat{\mathbf{m}}=\sigma(\mathbf{o})\in[0,1]^N.
\end{equation}

\subsection{Training Objective}\label{sec:training}

We adopt a decoupled training scheme: our point-consistent part encoder is trained with the contrastive objective~\cite{khosla2020supervised} in Sec.~\ref{sec:encoder}, while the scale-aware prompt decoder is optimized subsequently with the encoder frozen.

In the second stage, we randomly sample one target part per object, which is selected following the strategy of our experimental setup (Sec.~\ref{sec:comparison}).

However, each sample annotates only one part as positive, resulting in a systematically small within-sample positive ratio $\pi$ in this setting.
Using BCE alone tends to bias the optimization toward negatives/background and can admit a degenerate all-negative solution. To mitigate this bias while balancing set-level overlap and pointwise calibration, we employ a hybrid objective that combines Dice with a dynamically reweighted BCE. Given the predicted probability mask $\hat{\mathbf{m}}$ and ground truth $\mathbf{m}\in\{0,1\}^{N}$, the segmentation loss is
\begin{equation}
\mathcal{L}_{\text{seg}}
= \lambda_{\text{bce}}\,\mathrm{BCE}_{\text{dyn}}(\hat{\mathbf{m}},\mathbf{m})
\;+\;
\lambda_{\text{dice}}\Bigl(1-\tfrac{2\,\hat{\mathbf{m}}^\top\mathbf{m}}{\|\hat{\mathbf{m}}\|_1+\|\mathbf{m}\|_1}\Bigr).
\end{equation}
\input{fig/dataset_fig}
This design simultaneously optimizes pointwise probabilities and set-level overlap: $\mathrm{BCE}_{\text{dyn}}$ calibrates probabilities, whereas the Dice term directly improves set-level overlap and is more robust to class imbalance, thereby improving recall and maintaining gradient stability for small parts, long-tailed cases, and sparse masks. The dynamically reweighted BCE is defined as
\begin{equation}
\begin{aligned}
\mathrm{BCE}_{\text{dyn}}
&= -\frac{1}{N}\Bigl(\beta \sum_{j\in J_+}\log \hat{m}_j
          + \sum_{j\in J_-}\log(1-\hat{m}_j)\Bigr),\\
\beta &= \frac{1-\pi}{\pi+\varepsilon},\quad
\pi=\frac{1}{N}\sum_{j=1}^{N} m_j,\\
J_+ &= \{\,j \mid m_j=1\,\},\quad
J_-=\{\,j \mid m_j=0\,\},
\end{aligned}
\end{equation}
where $\beta$ is adaptively computed from the per-sample positive ratio $\pi$ to mitigate class imbalance, and $\varepsilon>0$ is a numerical stabilizer.

%% file: fig/dataset_fig.tex
\begin{figure*}[t]
    \centering
    \includegraphics[width=\textwidth]{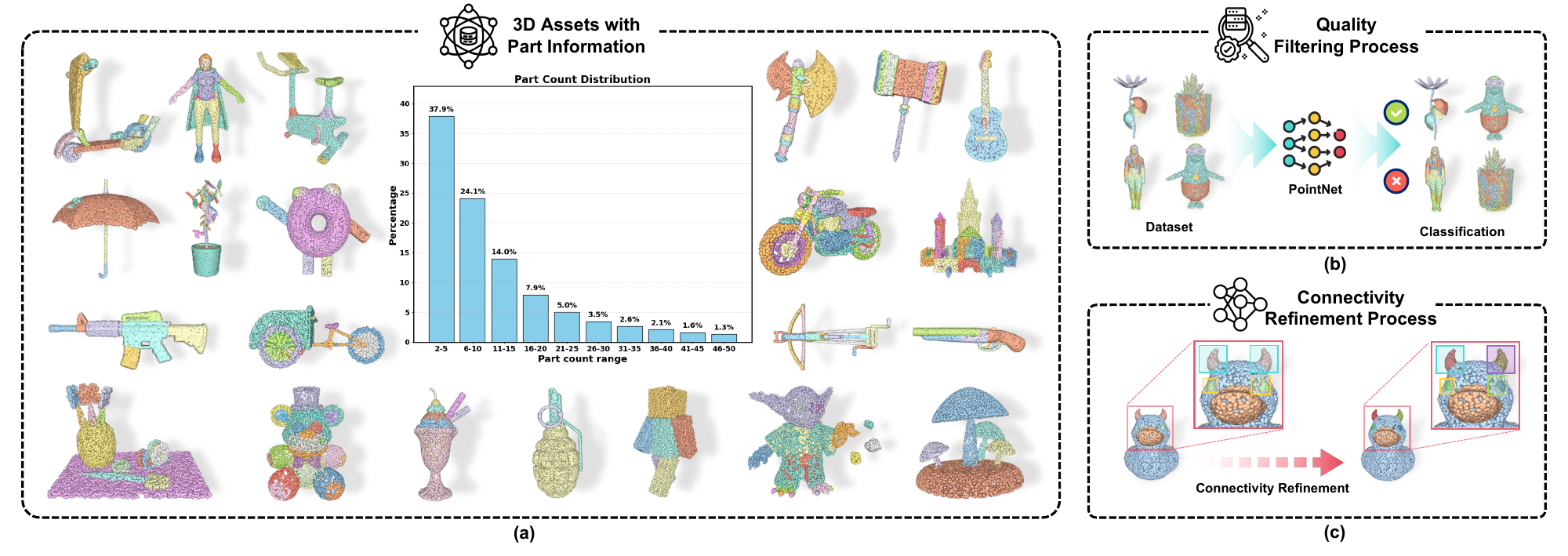}
    \caption{Dataset overview: covering diverse categories and providing high-quality part-level annotations; the histogram shows the long-tailed distribution of part counts.}
    \label{fig:dataset}
    \vspace{-1em}
\end{figure*}

%% file: sec/4_data.tex
\section{Dataset Curation}

To support 3D-supervised training, we introduce a scalable data pipeline for point cloud part segmentation labeling and curate a large-scale, high-quality labeled point cloud part segmentation dataset.
The pipeline comprises part annotation, quality filtering, and connectivity refinement to ensure robustness and diversity. We collect over 100,000 point cloud instances with around 1.2 million part labels, which, to the best of our knowledge, is one of the largest publicly available 3D part segmentation datasets.

\subsection{Scalable Data Pipeline}\label{sec:makedata}

The data pipeline consists of three steps: part annotation, quality filtering, and connectivity refinement. Given a 3D object, we adopt surface-area–proportional sampling and assign the corresponding part label to every sampled point, which follows the part-annotation mining strategy of PartCrafter~\cite{lin2025partcrafter}. However, such an automatic strategy inevitably introduces erroneous annotation artifacts, which may mislead the model training. To filter out possible wrong labels, we further propose quality filtering and connectivity refinement.

In the quality filtering stage (see Figure~\ref{fig:dataset}(b)), we first manually curate a small-scale dataset containing examples of both reasonable and unreasonable part annotations. We then train a binary PointNet~\cite{qi2017pointnet} validator on this dataset. Each point cloud is represented by $N$ points, each defined by a 4D vector $(x, y, z, l)$ encompassing normalized coordinates and its part-label index. Once trained, this validator is applied across the entire dataset to automatically screen and filter out samples with unreasonable annotations, ensuring that only high-quality labeled data proceed to subsequent stages.

We then perform connectivity refinement (see Figure~\ref{fig:dataset}(c)): artist annotations may group spatially disconnected yet semantically related regions under the same label, which degrades prompt-segmentation precision and hampers isolating local structures. For the points of each label, we compute the axis-aligned bounding-box diagonal length $d$ and set the DBSCAN~\cite{ester1996density} radius to $\epsilon = d \times \epsilon_{\text{factor}}$. If multiple spatial clusters are detected within a single label, we split it into distinct new labels. Finally, we filter and retain objects containing 2 to 50 parts to ensure dataset balance and suitability for part segmentation tasks.

\subsection{Dataset Overview}\label{sec:data}
We mainly collect 3D assets from Objaverse~\cite{deitke2023objaverse-xl,deitke2023objaverse}, which is one of the largest 3D object datasets. We filter more than 100,000 point cloud instances from this dataset, spanning over 400 categories. This category advantage lies in covering open-world shapes from simple objects to complex machinery, supporting broader geometric and semantic diversity. In total, we annotate around 1.2 million part labels, with an average of 11 parts per object. As shown in Figure~\ref{fig:dataset}(a), the part count distribution spans 2-50, exhibiting a long-tailed pattern to ensure balanced coverage of fine- and coarse-grained shapes. %Compared to existing datasets like PartNet~\cite{mo2019partnet}, our dataset is significantly larger both in instance scale (PartNet, around 26,000) and in the number of labeled parts (PartNet, 573,000). 
% and the long-tail distribution better captures rare complex structures. 
%In terms of quality, the screening process ensures label accuracy and geometric integrity. 

%% file: sec/5_experiment.tex
\section{Experiments}
\input{fig/exp_interactive_fig}
\subsection{Implementation Details}\label{sec:details}

We adopt a decoupled supervision strategy to separately train our proposed encoder and decoder. During training, we fix the tri-plane feature dimensions $D, H, W$ to $448, 512, 512$, respectively. The number of sampling points $N$ is set to 10,000.

\noindent\textbf{Encoder training.} The encoder is initialized with PartField~\cite{liu2025partfield} pre-trained parameters and then further optimized with contrastive learning on native point cloud annotations. For each mini-batch, we uniformly sample $\hat N=1000$ points to form $\hat P$ and compute the contrastive loss on it with temperature $\tau=0.07$. We use AdamW~\cite{loshchilov2017decoupled} with a learning rate of \(1\times 10^{-5}\). The encoder is trained for 15 epochs, which takes about one day on 8 NVIDIA A6000 GPUs.

\noindent\textbf{Decoder training.}
In this phase, we freeze the parameters of the encoder and fine-tune the decoder only. We set the layer number of scale modulation \(L_m\) to 2 and the layer number of bi-directional cross-attention blocks \(L_d\) to 4. The segmentation loss is a weighted sum of dynamic BCE and Dice, with $\lambda_{\mathrm{bce}}=0.7$ and $\lambda_{\mathrm{dice}}=0.3$. During training, we apply the scale dropout with a probability of 0.1, and at inference, we threshold the probability at \(\theta=0.7\). We use AdamW~\cite{loshchilov2017decoupled} with a learning rate of \(1\times 10^{-4}\). The training batch size is 40. The decoder is trained for 110 epochs with 4 NVIDIA A6000 GPUs.

\noindent\textbf{Training data.} We train both the encoder and decoder on our curated dataset (Sec.~\ref{sec:data}) and PartNet, removing PartNet instances whose IDs appear in PartNet-E to prevent data leakage.
\subsection{Comparison}\label{sec:comparison}
\input{tab/comparison1}

% \noindent\textbf{Task Setup.}
We compare S$^2$AM3D with state-of-the-art methods on two tasks: \emph{(i) Interactive Segmentation} and \emph{(ii) Full Segmentation}. Interactive segmentation focuses on generating a single segment corresponding to the prompt point, while full segmentation requires predicting part labels for all the points. These two tasks demand a robust perception of both local point coherence and global semantic modeling. 
% In our full-segmentation evaluation, a point prompt is provided to the model for each ground-truth part to generate its mask. These masks are then consolidated to yield a comprehensive segmentation for the entire object. Please refer to Suppl. for more information.

\input{fig/exp_fullseg_fig}
\input{tab/comparison2}
\noindent\textbf{Evaluation Datasets.}
We choose PartObjaverse-Tiny~\cite{yang2024sampart3d} and PartNet\mbox{-}E~\cite{liu2023partslip} for experimental evaluation.
PartObjaverse-Tiny~\cite{yang2024sampart3d} is a compact subset of Objaverse~\cite{deitke2023objaverse,deitke2023objaverse-xl}, containing 200 samples across 8 categories with manual part annotations.
PartNet\mbox{-}E~\cite{liu2023partslip}
provides 1,906 point cloud shapes across 45 categories, which can be used to assess cross-category generalization.

\noindent\textbf{Baselines.}
For interactive segmentation, we compare two point-prompt baselines: Point-SAM~\cite{zhou2024point} and P\textsuperscript{3}-SAM~\cite{ma2025p3}. For fair comparison, all methods are evaluated with their default input configurations: Point-SAM~\cite{zhou2024point} employs both coordinates and color information, while P\textsuperscript{3}-SAM~\cite{ma2025p3} utilizes coordinates and surface normals. In contrast, our approach requires only XYZ coordinates.
For full segmentation, we select current state-of-the-art methods, including Find3D~\cite{ma2025find}, SAMPart3D~\cite{yang2024sampart3d}, SAMesh~\cite{tang2024segment}, PartField~\cite{liu2025partfield}, and P\textsuperscript{3}-SAM~\cite{ma2025p3}.
% In our full-segmentation evaluation, a point prompt is provided to the model for each ground-truth part to generate its mask. These masks are then consolidated to yield a comprehensive segmentation for the entire object. 
% Since artist-authored meshes in PartObjaverse-Tiny may contain strong connectivity cues aligned with ground-truth part labels, we evaluate mesh-based methods following the \emph{watertight-mesh} protocol~\cite{ma2025p3}. 
% Please refer to Suppl. for more information.

\noindent\textbf{Metrics.} 
We use Intersection over Union (IoU) as the evaluation metric. 
% For each object, IoU is computed averaged over all of its parts to obtain a per-sample score, which is then averaged across the dataset. 
The metric is computed across all parts of a single object to obtain the per-sample score, which is subsequently averaged over the entire dataset.
 For interactive segmentation, we follow the strategy in Point-SAM~\cite{zhou2024point}, which selects the prompt point for each part as the interior point farthest from its boundary. All methods are evaluated on 10,000 points normalized to the unit sphere. For full segmentation, we follow the experimental protocol in P\textsuperscript{3}-SAM~\cite{ma2025p3}. For our method, a point prompt is provided to the model for each ground-truth part to generate its mask. These masks are then consolidated to yield a comprehensive segmentation for the entire object. Please refer to Suppl. for more information.
% and report the Intersection-over-Union (IoU) across all objects in the dataset.

\noindent\textbf{Results.} 
% For interactive part segmentation, we evaluate S\textsuperscript{2}AM3D with scale enabled and disabled to quantify performance when scale information is unavailable.
We report quantitative results of interactive segmentation in Table~\ref{tab:inter_exp}. Although our method can optionally incorporate scale prompts to boost performance, we also provide a non-scale version to ensure a fair comparison with the other two methods~\cite{zhou2024point,ma2025p3}. 
Table~\ref{tab:inter_exp} shows that S$^2$AM3D achieves the best results in both datasets~\cite{yang2024sampart3d,liu2023partslip}. With the input of scale prompts, our method further achieves a performance gain of 14.72\% and 14.99\%, demonstrating the effectiveness of our scale-aware decoding. 
The qualitative results in Figure~\ref{fig:exp_interactive} are basically consistent with the quantitative results. Our model produces segmentations largely consistent with the GT without scale condition; supplying scale guidance corrects the few cases of granularity mismatch by calibrating the level of detail.
% With scale, our average improvements over Point-SAM and P\textsuperscript{3}-SAM are approximately 52\% and 66\%; without scale, they remain about 18\% and 28\%.
% In particular, P\textsuperscript{3}-SAM is originally trained/evaluated with 100,000 points; when unified with 10,000 points, its performance decreases substantially, indicating a strong dependence on high-density point clouds. Qualitatively, Fig.~\ref{fig:exp_interactive} shows that with a single click, our predictions exhibit cleaner boundaries, a more precise response to the target, and robust performance across targets of varying granularity;
% Fig.~\ref{fig:exp_scale} further demonstrates smooth, controllable transitions from coarse to fine segmentation as the continuous scale prompt varies, enabled by our scale-aware decoder aligning segmentation granularity with user intent.

For full segmentation, Table~\ref{tab:full_seg} shows that S$^2$AM3D achieves 63.29 mIoU on PartObjaverse-Tiny and 77.98 mIoU on PartNet\text{-}E~\cite{liu2023partslip}. We also visualize the segmentation results in Figure~\ref{fig:exp_fullseg}. 
% that the baseline methods exhibit several typical limitations: 
Constrained by 2D priors, methods like PartField~\cite{liu2025partfield} and SAMPart3D~\cite{yang2024sampart3d} produce segmentations with poor 3D consistency. On the other hand, 3D native methods like SAMesh~\cite{tang2024segment} and P\textsuperscript{3}-SAM~\cite{ma2025p3} tend to fail in some long-tailed cases.
% tends to over-segment, whereas P\textsuperscript{3}-SAM yields fragmented and incomplete masks due to their inherent design~\cite{tang2024segment,ma2025p3}.
In contrast, S$^2$AM3D effectively circumvents these issues, achieving segmentation with higher completeness and better consistency on complex structures.

\input{fig/ablation_fig}
\input{fig/exp_scale_fig}

\subsection{Ablation Studies}\label{sec:ablation}

To demonstrate the effectiveness of our proposed modules and data, we conduct ablation studies in Table~\ref{tab:ablation}.
% reports three ablations that probe native 3D refinement, our curated data, and scale conditioning. Except for the removal of specified components, all training and inference protocols are kept identical across variants.

% \noindent\textbf{3D Contrastive Supervision.} Replacing the encoder with a PartField-pretrained model produces the largest performance degradation among all ablation variants, showing that encoder-side 3D refinement is the principal contributor to the gains. The main cause of this performance gap is visually demonstrated in Fig.~\ref{fig:ablation}: compared to the clear, coherent features produced with native 3D supervision, features extracted by the ablated model exhibit noticeable blurry boundaries and internal inconsistencies. This difference in feature quality directly translates to contrasting decoding outcomes: the full model produces coherent segmentation masks, while the ablated model yields fragmented results. This clearly demonstrates that improved feature consistency is crucial for achieving high-quality segmentation.
\noindent\textbf{3D Contrastive Supervision.} Among all ablation variants, replacing an encoder without 3D supervision yields the most significant performance degradation. This finding unequivocally establishes that encoder-side 3D feature refinement is the principal contributor to the observed gains. The rationale behind this substantial performance gap is vividly illustrated in Figure~\ref{fig:ablation}: features extracted by the ablated model exhibit noticeable blurry boundaries and internal inconsistencies compared to our pre-trained encoder. This disparity in feature quality directly translates to contrasting decoding outcomes: the full model generates coherent and accurate segmentation masks, whereas the ablated model produces fragmented and inconsistent results. This analysis clearly demonstrates that achieving improved feature consistency and geometric fidelity is paramount for high-quality point cloud segmentation.

% \noindent\textbf{Our curated data.} Restricting training to the PartNet split (i.e., removing Ours-Data) produces consistent drops across all benchmarks ~\cite{mo2019partnet}. This suggests that, after systematic filtering and annotation governance, the curated data provides a distributional complement in terms of geometric variation and part composition, translating into stable cross-dataset gains.
\noindent\textbf{Our Curated Data.} To verify the impact of our own data, we train our model with PartNet~\cite{mo2019partnet} dataset, which results in performance drops compared to the model trained with our data. This finding strongly indicates that our curated data, having undergone systematic filtering and annotation governance, provides a crucial distributional complement to existing benchmarks. Specifically, it introduces richer geometric variation and diverse part composition, which directly translates into stable cross-dataset generalization gains for our model.

\noindent\textbf{Scale conditioning.} 
% In this ablation the entire scale pathway is removed: no scale signal \(s\) is provided during training or testing, and all \(s\)-dependent modules (learnable sinusoidal embedding and FiLM) are disabled; results are therefore reported only under the No scale protocol. The gap to the Scale provided–Full setting establishes the contribution of explicit scale when available. Moreover, within No scale, the No scale–Full variant (keeping the pathway but not injecting \(s\)) consistently outperforms w/o Scale embedding, indicating that the associated architectural inductive bias confers a modest benefit even without scale input.
We visualize the effectiveness of scale prompting in Figure~\ref{fig:exp_scale}. The results show that our method can provide reasonable segmentation without scale information, while scale conditioning allows more flexible and controllable segmentation for users.
To further quantify the contribution of explicit scale information, we train a model without our scale embeddings. As shown in the last row of Table~\ref{tab:ablation}, even though the scale condition is not provided, our model still outperforms the one without scale embeddings. This comparison indicates that scale embeddings enhance the robustness of feature decoding, benefiting the training of our proposed decoder.

\input{tab/ablation1}

%% file: fig/exp_interactive_fig.tex
\begin{figure*}[t]
    \centering
    \includegraphics[width=\textwidth]{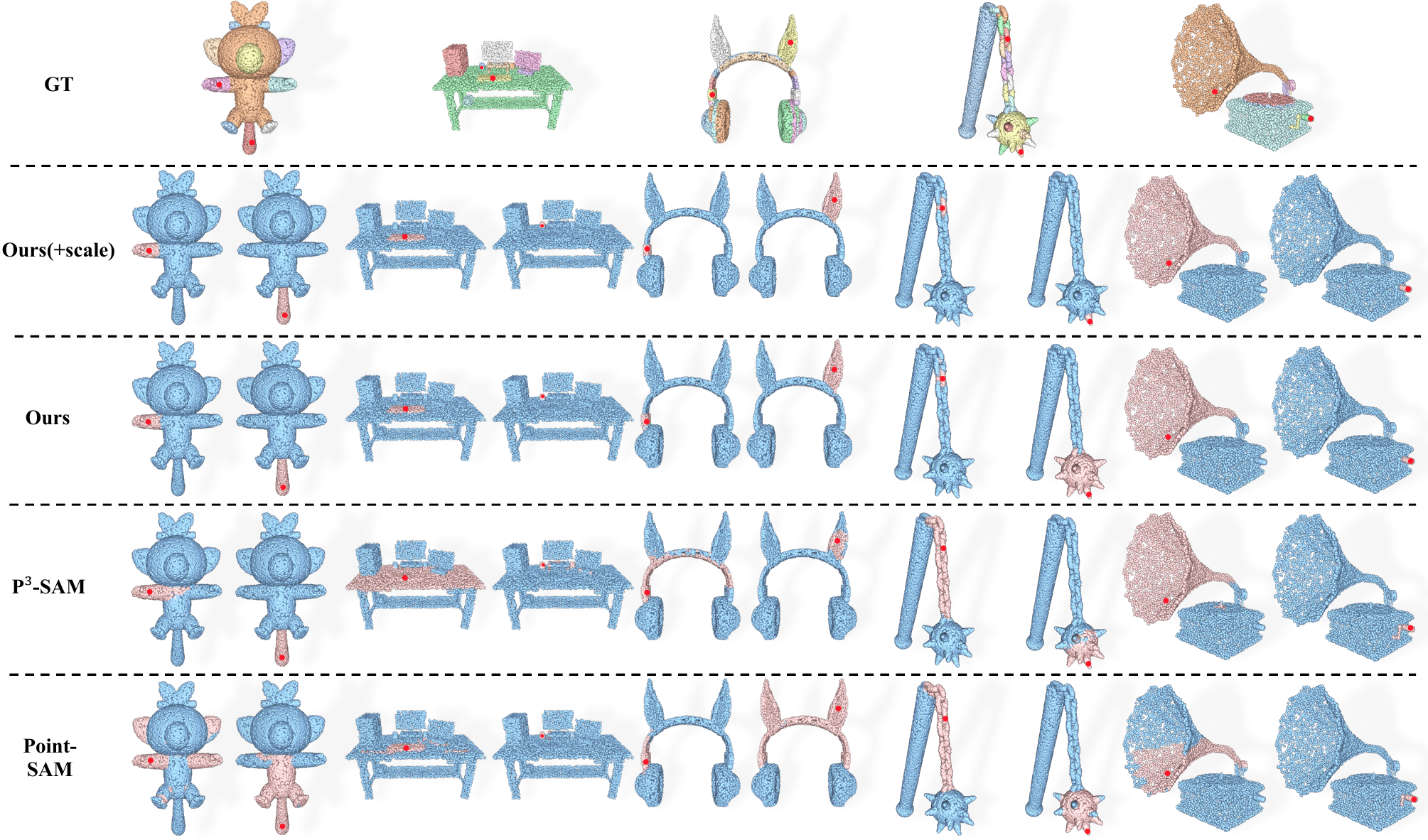}
    \caption{Qualitative comparison on our curated dataset (see Sec.~\ref{sec:data}). With a point prompt, S\textsuperscript{2}AM3D responds more accurately to the target, producing masks with cleaner boundaries and more complete topology.}
    \label{fig:exp_interactive}
    \vspace{-0.5em}
\end{figure*}

%% file: tab/comparison1.tex
\begin{table}[t]
\centering
\small
\caption{Quantitative comparison of interactive part-level segmentation. We report IoU (\%) on PartObjaverse-Tiny and PartNet-E. We also provide a scale-aware version \texttt{(+scale)}, which has an additional scale prompt. The best score in each column is highlighted in \textbf{bold}, and \underline{underline} denotes the second best.}
% \textit{Avg.} is the mean of PartObjaverse-Tiny and PartNet\text{-}E.}
\label{tab:inter_exp}
\begin{tabular}{l c c c}
\toprule
Method & PartObjaverse-Tiny & PartNet\text{-}E & \textit{Avg.} \\
\midrule
Point\text{-}SAM~\cite{zhou2024point}        & 31.46 & 50.23 & 40.85 \\
P\textsuperscript{3}-SAM~\cite{ma2025p3}           & 35.05 & 39.98 & 37.52 \\
Ours        & \underline{46.47} & \underline{62.52} & \underline{54.50} \\
Ours (+scale)           & \textbf{61.19} & \textbf{77.51} & \textbf{69.35} \\
\bottomrule
\end{tabular}
\vspace{-0.5em}
\end{table}

%% file: fig/exp_fullseg_fig.tex
\begin{figure*}[t]
    \centering
    \includegraphics[width=\textwidth]{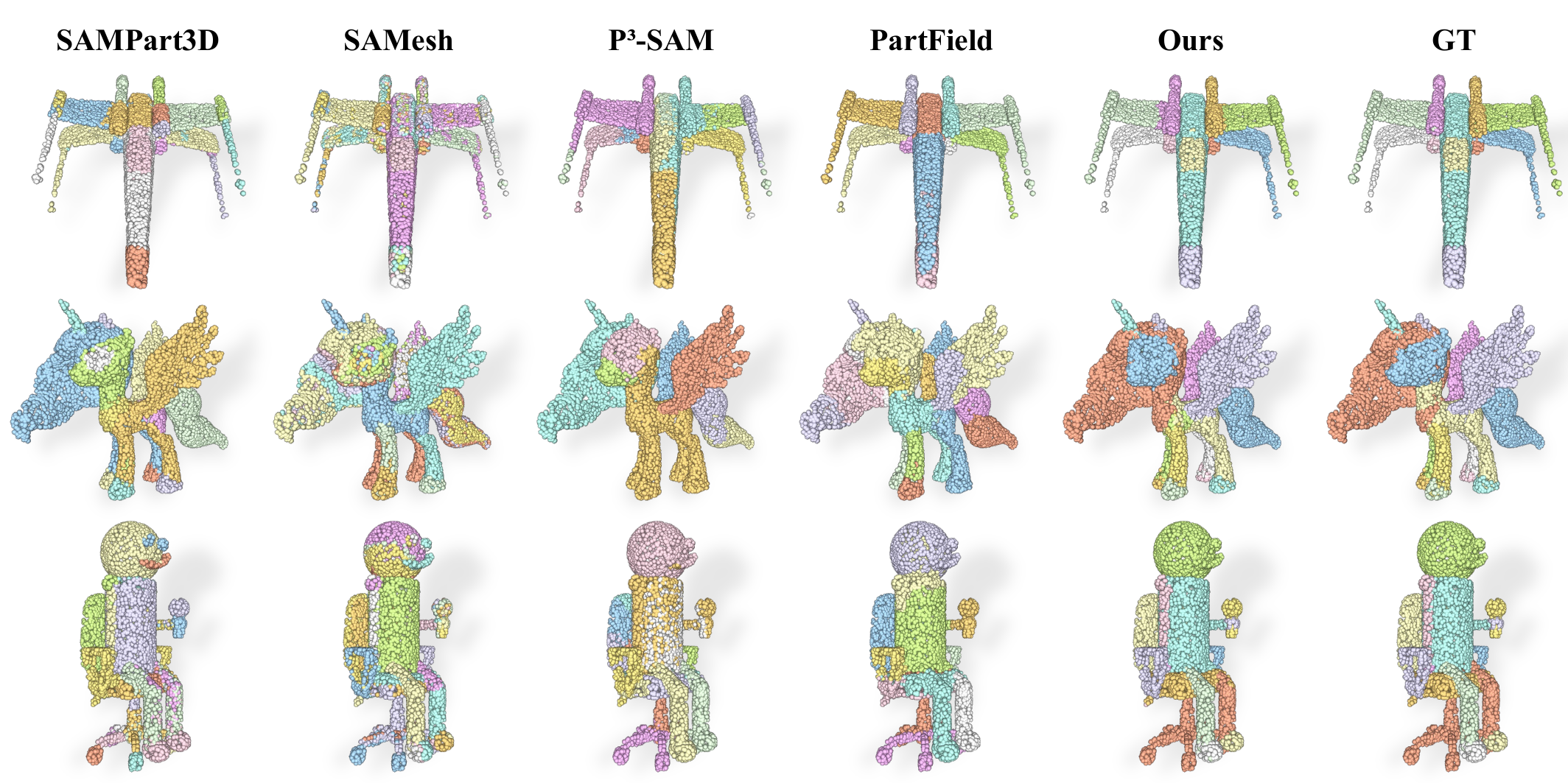}
    \caption{Qualitative comparison of full segmentation (PartObjaverse-Tiny~\cite{yang2024sampart3d}). For ease of comparison with our point cloud method, mesh-level outputs are presented as point clouds by uniformly sampling the segmented meshes.}
    \label{fig:exp_fullseg}
    \vspace{-0.5em}
\end{figure*}

%% file: tab/comparison2.tex
\begin{table}[t]
\centering
\small
\setlength{\tabcolsep}{6pt}
\caption{Quantitative comparison of full segmentation.}
% We report mean IoU (\%) on instance-level labels. The best score in each column is highlighted in \textbf{bold}. \textit{Avg.} is the mean of PartObjaverse-Tiny and PartNet\text{-}E.}
\label{tab:full_seg}
\begin{tabular}{lccc}
\toprule
Method & PartObjaverse-Tiny & PartNet-E & \textit{Avg.} \\
\midrule
Find3D~\cite{ma2025find}        & 20.76 & 21.69 & 21.23 \\
SAMPart3D~\cite{yang2024sampart3d}     & 48.79 & 56.17 & 52.48 \\
SAMesh~\cite{tang2024segment}        & - & 26.66 & - \\
PartField~\cite{liu2025partfield}     & 51.54 & 59.10 & 55.32 \\
P\textsuperscript{3}-SAM~\cite{ma2025p3} & 58.10 & 65.39 & 61.75 \\
Ours          & \textbf{63.29} & \textbf{77.98} & \textbf{70.64} \\
\bottomrule
\end{tabular}
\vspace{-0.5em}
\end{table}

%% file: fig/ablation_fig.tex
\begin{figure}[t]
    \centering
    \includegraphics[width=0.45\textwidth]{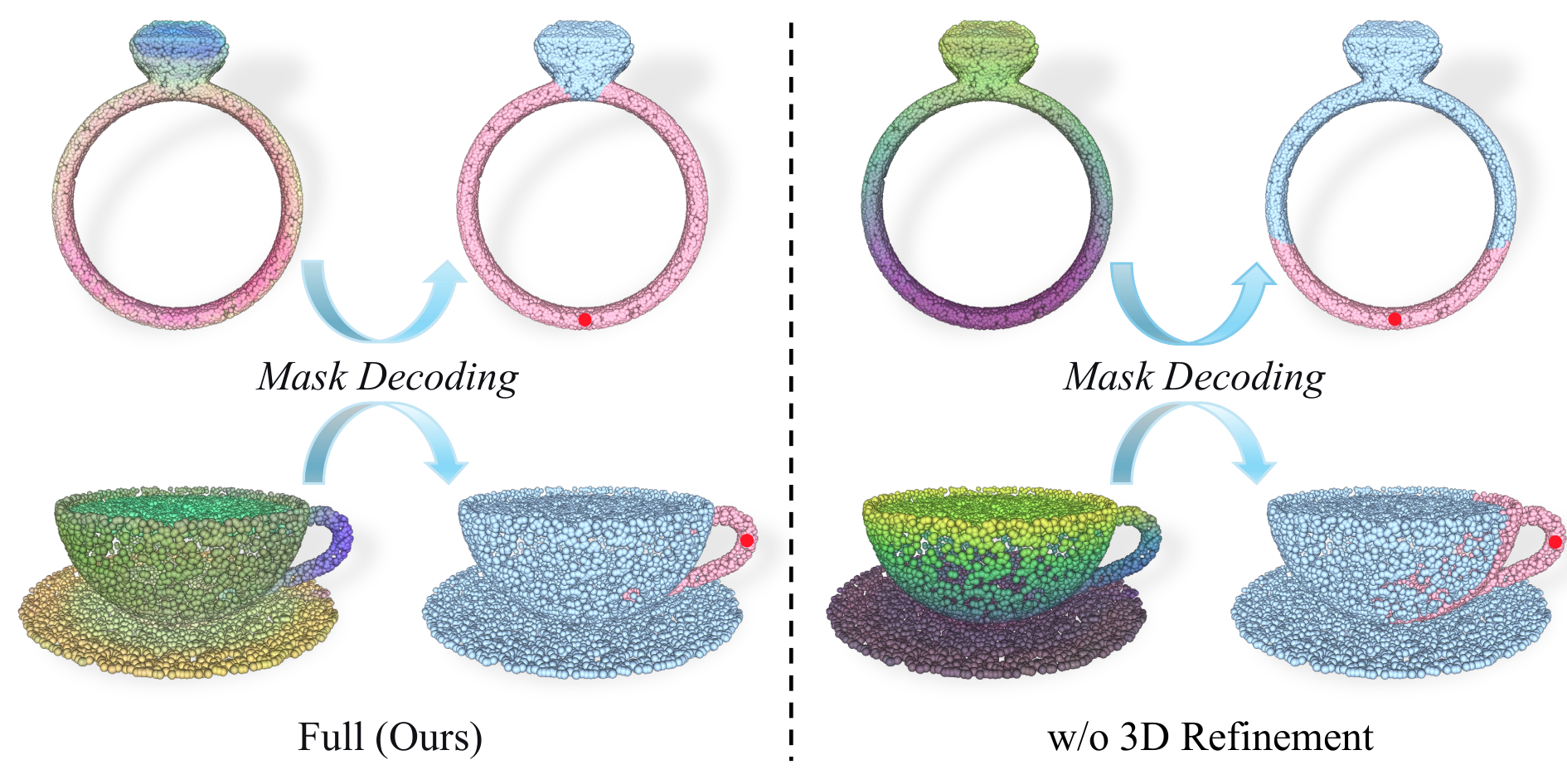}
    \caption{Visualization of the ablation study on encoder feature quality. This side-by-side comparison displays extracted features and their resulting segmentations for our full model (left) and the ablated model without 3D refinement (right).}
    \label{fig:ablation}
    \vspace{-1em}
\end{figure}

%% file: fig/exp_scale_fig.tex
\begin{figure}[t]
    \centering
    \includegraphics[width=0.45\textwidth]{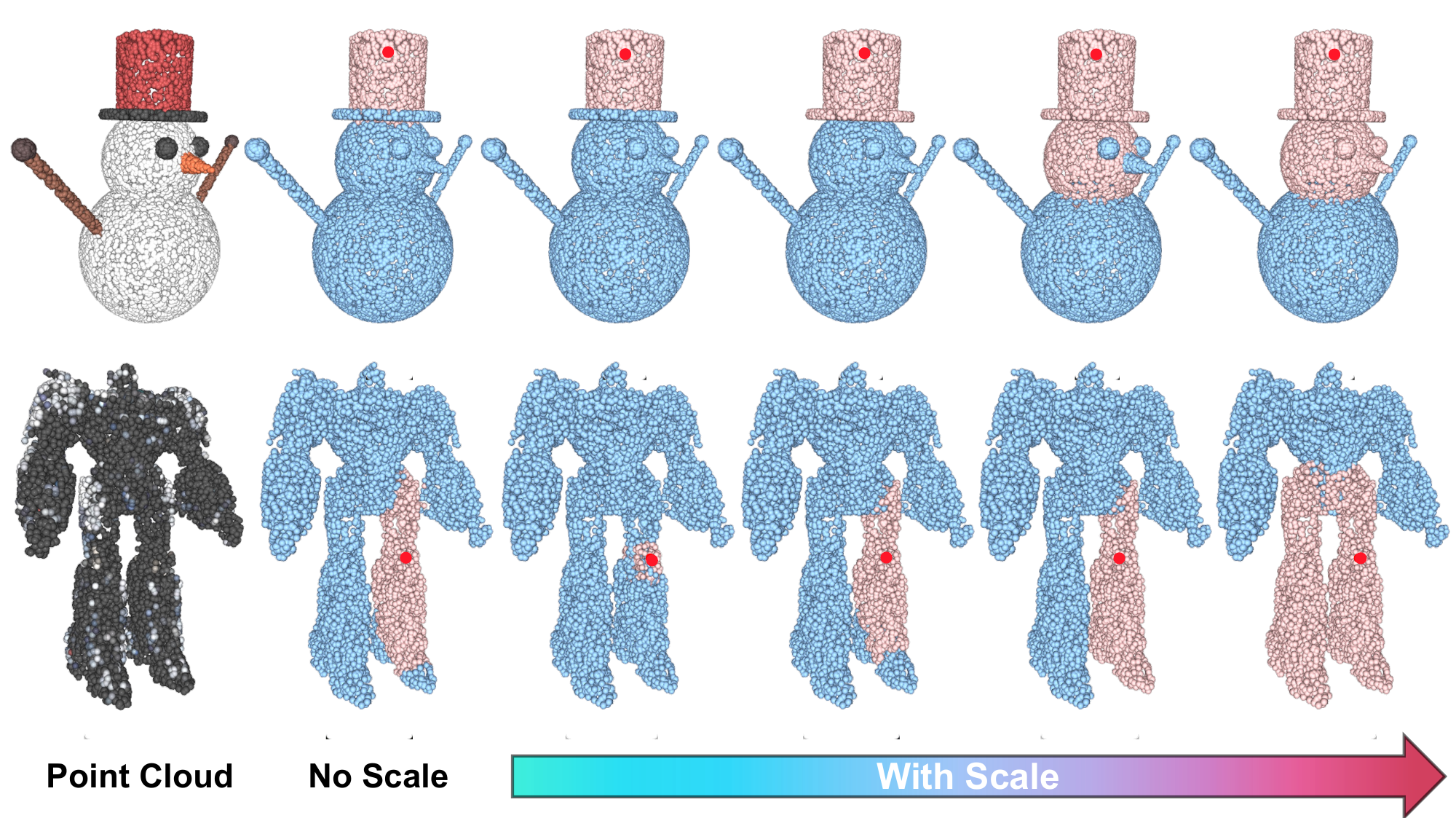}
    \caption{Visualization of continuous scale controllability. With the same prompt point, as the scale $s$ increases from 0 to 1, the segmentation transitions smoothly from fine to coarse; a \texttt{(No scale)} counterpart is also provided for reference.}
    \label{fig:exp_scale}
    \vspace{-1em}
\end{figure}

%% file: tab/ablation1.tex
\begin{table}[t]
\centering
\small
\setlength{\tabcolsep}{4pt}
\begin{tabular}{lccc}
\toprule
\textbf{Setting} & PartObjaverse-Tiny & PartNet-E & Avg. \\
\midrule
\multicolumn{4}{l}{\textbf{+scale}} \\
Full                   & \textbf{61.19} & \textbf{77.51} & \textbf{69.35} \\
w/o 3D supervision          & 53.94 & 64.11 & 59.03 \\
w/o our data          & 53.12 & 66.12 & 59.62 \\
\midrule
\multicolumn{4}{l}{\textbf{No scale}} \\
Full                   & \textbf{46.47} & \textbf{62.52} & \textbf{54.50} \\
w/o 3D supervision          & 41.14 & 55.39 & 48.27 \\
w/o our data          & 42.12 & 58.56 & 50.34 \\
w/o scale embedding    & 42.31 & 58.28 & 50.30 \\
\bottomrule
\end{tabular}
\caption{Ablation studies (mIoU, \%). Groups indicate whether scale information is provided at test time: \textbf{+scale} means a scale condition is given, while \textbf{No scale} means it is not.}
\label{tab:ablation}
\vspace{-1em}
\end{table}

%% file: sec/6_conclusion.tex
\section{Conclusion}

We present \textbf{S$^2$AM3D}, a scale-controllable framework for part segmentation of 3D point clouds. Through the design of a \textit{point-consistent encoder} and a \textit{scale-aware prompt decoder}, our method addresses the challenge of cross-view inconsistency, enhances global feature consistency, and enables real-time adjustment of segmentation granularity via continuous scale signals. The construction of a large-scale, high-quality part-level point cloud dataset provides crucial support for model training. Extensive experiments demonstrate that S$^2$AM3D achieves leading performance on multiple benchmarks, exhibiting robustness and controllability when handling complex structures and multi-scale parts, thereby providing a reliable solution for fine-grained 3D scene understanding and content editing. While the current framework primarily relies on point prompts and scale signals for interaction, future work will explore incorporating richer prompt modalities such as text instructions to support more intuitive semantic interaction.

\section*{Acknowledgements}
This work was supported by the National Key R\&D Program of China under Grant No.~2022YFA1004100 and the National Natural Science Foundation of China (NSFC) under Grant No.~62476067.\par\par

%% file: tab/suppl_density_tab.tex
\begin{table}[t]
\centering
\small
\begin{tabular}{l cc cc}
\toprule
Method &
\multicolumn{2}{c}{PartObjaverse-Tiny} &
\multicolumn{2}{c}{PartNet\text{-}E} \\
& 10k & 100k & 10k & 100k \\
\midrule
Ours          & 46.47 & 46.46 & 62.52 & 62.41 \\
Ours (+scale) & 61.19 & 60.98 & 77.51 & 77.67 \\
\bottomrule
\end{tabular}
\caption{Effect of point density on S$^2$AM3D in the interactive setting.}
\label{tab:density_robust}
\end{table}

%% file: tab/suppl_complex_tab.tex
\begin{table}[t]
\centering
\begin{tabular}{lcc}
\toprule
Method & Params (M) & Time (ms) \\
\midrule
Point-SAM~\cite{zhou2024point} & 311 & $\sim$5 \\
P$^3$-SAM~\cite{ma2025p3}   & 112 & $\sim$3 \\
Ours             & 120 & $\sim$3 \\
\bottomrule
\end{tabular}
\caption{Complexity analysis of different methods.}
\label{tab:complexity}
\end{table}

%% file: tab/suppl_scale_tab.tex
\begin{table*}[t]
\centering
\begin{tabular}{c|rrrrrrrr}
\toprule
$|\delta|$ & $5\%$ & $10\%$ & $20\%$ & $30\%$ & $50\%$ & $100\%$ & $200\%$ & $300\%$ \\
\midrule
$\Delta$IoU ($-\delta$) & $+0.15$ & $+0.12$ & $-0.05$ & $-0.66$ & $-2.99$ & $-7.43$ & $-14.85$ & $-20.19$ \\
$\Delta$IoU ($+\delta$) & $-0.19$ & $-0.38$ & $-0.78$ & $-1.48$ & $-3.56$ & $-7.47$ & $-14.17$ & $-18.98$ \\
\bottomrule
\end{tabular}
\caption{Effect of relative scale perturbation on mean IoU on PartObjaverse-Tiny~\cite{yang2024sampart3d}. $\Delta$IoU is measured with respect to the $\delta=0$ setting.}
\label{tab:scale_robustness}
\end{table*}

%% file: fig/more_vis_fig.tex
\begin{figure*}[t]
    \centering
    \includegraphics[width=\textwidth]{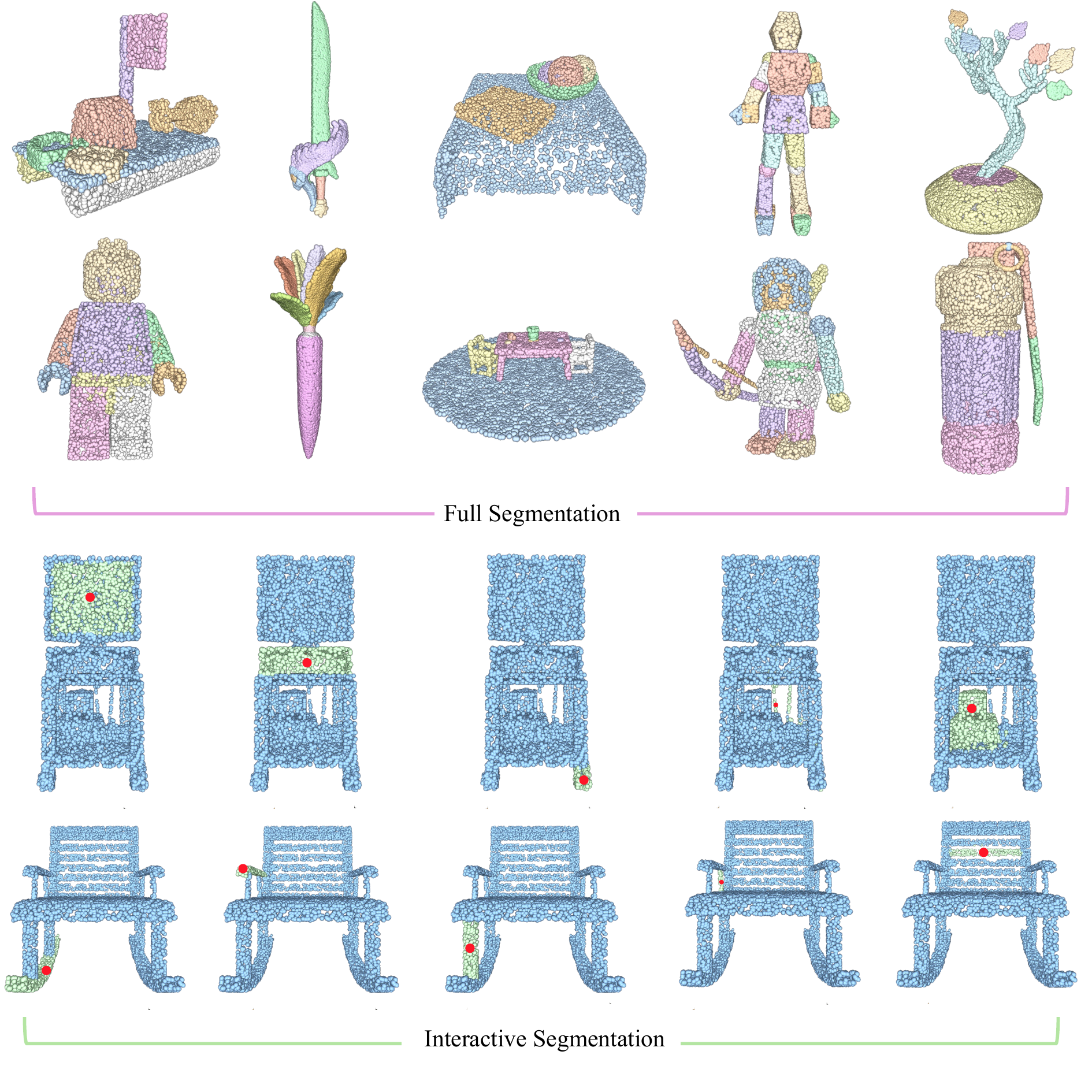}
\caption{Additional qualitative results of S$^2$AM3D on full segmentation and interactive segmentation on our curated dataset.}

    \label{fig:more_vis}
\end{figure*}